%% file: acl.tex
\pdfoutput=1

\documentclass[11pt]{article}

\usepackage[final]{acl}

\usepackage{times}
\usepackage{latexsym}
\usepackage{cleveref}
\usepackage{booktabs}
\usepackage{multirow}
\usepackage[T1]{fontenc}
\usepackage[utf8]{inputenc}
\usepackage{microtype}
\usepackage{inconsolata}
\usepackage{amssymb}
\usepackage{pifont}
\usepackage[framemethod=TikZ]{mdframed}
\usepackage{graphicx}
\usepackage{subcaption}
\usepackage{enumitem}

\usepackage[T1]{fontenc}

\usepackage[utf8]{inputenc}

\usepackage{microtype}

\usepackage{inconsolata}

\newmdenv[%
    backgroundcolor=gray!10,
    linecolor=gray!70,
    linewidth=1pt,
    roundcorner=5pt,
    skipabove=10pt,
    skipbelow=10pt,
    font=\ttfamily\small,
]{promptbox}

\newmdenv[
    backgroundcolor=blue!5,
    linecolor=blue!70,
    linewidth=1pt,
    roundcorner=5pt,
    skipabove=10pt,
    skipbelow=10pt,
    font=\ttfamily\small,
]{keytakeaway}

\title{An Empirical Study of In-context Learning \\ in LLMs for Machine Translation}

\input{content/authors}

\begin{document}
\maketitle
\input{content/abstract}

\input{content/introduction}
\input{content/related_works}
\input{content/methodology}
\input{content/experiment_setup}
\input{content/results_and_analysis}
\input{content/conclusion}
\input{content/limitations}
\input{content/ethics}
\input{content/acknowledgements}
\bibliography{anthology,custom}
\clearpage

\appendix
\input{appendix/A}

\end{document}

%% file: content/authors.tex
\author{
  \textbf{Pranjal A. Chitale$^{1,2}$}\thanks{~~Equal contribution}~~ \quad \textbf{Jay Gala~$^{3*}$\thanks{\;Work done at Nilekani Centre at AI4Bharat}} \quad \textbf{Raj Dabre$^{4}$}
  \\[1ex]
  $^{1}$Nilekani Centre at AI4Bharat \quad $^{2}$IIT Madras \\ $^{3}$Mohamed bin Zayed University of Artificial Intelligence \\
  $^{4}$National Institute of Information and Communications Technology, Kyoto, Japan
  \\[1ex]
    \texttt{cs21s022@cse.iitm.ac.in, jay.gala@mbzuai.ac.ae, prajdabre@nict.go.jp}
}

%% file: content/abstract.tex
\begin{abstract}
Recent interest has surged in employing Large Language Models (LLMs) for machine translation (MT) via in-context learning (ICL) \citep{vilar-etal-2023-prompting}. Most prior studies primarily focus on optimizing translation quality, with limited attention to understanding the specific aspects of ICL that influence the said quality. To this end, we perform the first of its kind, an exhaustive study of in-context learning for machine translation. We first establish that ICL is primarily example-driven and not instruction-driven. Following this, we conduct an extensive exploration of various aspects of the examples to understand their influence on downstream performance. Our analysis includes factors such as quality and quantity of demonstrations, spatial proximity, and source versus target originality. Further, we also investigate challenging scenarios involving indirectness and misalignment of examples to understand the limits of ICL. While we establish the significance of the quality of the target distribution over the source distribution of demonstrations, we further observe that perturbations sometimes act as regularizers, resulting in performance improvements. Surprisingly, ICL does not necessitate examples from the same task, and a related task with the same target distribution proves sufficient. We hope that our study acts as a guiding resource for considerations in utilizing ICL for MT. Our code is available on \url{https://github.com/PranjalChitale/in-context-mt-analysis}.
\end{abstract}

%% file: content/introduction.tex
\section{Introduction}
Large Language Models leverage In-Context Learning to effectively solve diverse downstream tasks \citep{brown2020language,dong2023survey,liuiclsurvey2023,openai2023gpt4,chowdhery2022palm}. This inference-only method involves conditioning the model with task-specific demonstrations consisting of input-output pairs within the prompt before the actual test example. Most recently, MT via ICL has become popular \cite{vilar-etal-2023-prompting}, however, there is still a limited understanding of how ICL works and its aspects in this context. \citet{raunak-etal-2023-dissecting} make some initial explorations on partial test sets, but there are still open questions pertaining to whether ICL is example-driven or instruction-driven, whether all examples contribute equally or not, and its potential to enforce control over the generation to regulate the pre-training biases observed in models.
To answer these questions, in this paper, we conduct experiments by perturbing either the instructions or the demonstrations while keeping the other clean to assess the model's ability to perform tasks appropriately. This allows us to understand if clear instructions can guide the model effectively even when examples are perturbed and, conversely, if clean examples can steer the model to perform the appropriate task despite the instruction being misleading.
We assess 6 models across 2 language families, spanning 12 language pairs on complete test sets across various noise thresholds by considering realistic perturbation attacks. 

When perturbing examples, we study  1) whether ICL treats all examples equally or favors those in spatial proximity to the test example, 2) whether utilizing in-context examples from related tasks can effectively guide models in performing another task to simulate the unavailability of demonstrations for some language pairs, 3) whether various choices of auxiliary source language for the demonstrations influence downstream MT, 4) whether models can implicitly make associations based on contextual information via a transitive MT setup and finally, 5) whether smaller models are susceptible to contextual misinformation or if their pre-training biases act as safeguards against it.
Our work aims to holistically understand ICL and its driving factors for MT tasks with experiments spanning over \textit{25K} runs across various models and language directions. We observe that examples significantly influence ICL, while instructions have a limited impact. Notably, target distribution plays a more significant role than source distribution in ICL demonstrations. While perturbations might seem potentially harmful, in some cases, they can act as a form of regularization, particularly evident in the Llama 2 7B model. Spatial proximity emerges as a critical factor, implying that clean examples should be placed closer and noisy ones farther during in-context example selection for optimal downstream performance. Furthermore, our findings suggest that examples from related tasks can suffice for ICL in MT, with the constraint that the target language of the demonstrations matches the language of the test example while the choice of the source language is inconsequential. The directionality of the demonstrations has minimal impact, with both target-original and source-original demonstrations being equally effective. ICL has the potential to override semantic priors and may be exploited to misguide models or generate misinformation. We believe our study offers valuable insights for practitioners and would be widely generalizable.

%% file: content/related_works.tex
\section{Related Works}

\noindent \textbf{In-context Learning:}
Large Language Models have demonstrated strong performance in various downstream tasks through supervised fine-tuning on task-specific data \citep{devlin-etal-2019-bert,raffel-etal-2020-t5,lewis-etal-2020-bart,radford2019language}. ICL \citep{brown2020language,dong2023survey,liuiclsurvey2023,openai2023gpt4,chowdhery2022palm} is a training-free method that has emerged as a cost-effective approach given the increasing size of LLMs ($\geq$ 7B parameters). 
ICL aims to implicitly learn the task-solving abilities through a few hand-crafted demonstrations. These emergent capabilities \citep{wei2022emergent,lu2023emergent} are driven by the distribution of the pretraining data \citep{chan2022data,briakou-etal-2023-searching}. 
Intuitively, ICL can be considered analogous to performing a gradient descent implicitly during the inference \citep{urek2023what,li2023transformers,zhang2023trained}.
\citet{vilar-etal-2023-prompting} explored few-shot prompting strategies for MT task on PaLM \citep{chowdhery2022palm} suggesting that the downstream MT performance is largely correlated with the quality of demonstrations. Subsequently, several recent works have also explored ICL capabilities of LLMs for the MT task \cite{robinson-etal-2023-chatgpt,10.5555/3618408.3620130,zhu2023multilingual,m-etal-2023-ctqscorer}. However, none of these works delve into a fine-grained examination of the underlying factors influencing the model performance; a gap that we fill with our study.

\noindent \textbf{Factors impacting in-context learning:}
Several aspects of the demonstrations, such as input distribution, output distribution, or input-output alignment, can play a significant role in the ICL performance of LLMs. While \citet{min-etal-2022-rethinking} observed limited significance of label correctness in demonstrations for classification tasks, however, \citet{yoo-etal-2022-ground,kossen2023incontext} argued that ICL is sensitive to input-label mapping and can affect downstream performance. Furthermore, \citet{wei2023larger} showed that ICL can override semantic priors, being influenced even by semantically unrelated input-label mappings, particularly in larger models. While prior works \citep{min-etal-2022-rethinking,wei2023larger,kossen2023incontext} have explored various aspects of ICL for NLU tasks, it is crucial to examine its applicability to NLG tasks. \citet{raunak-etal-2023-dissecting}, which is most related to our work, found that syntactic perturbations to demonstrations affect MT performance, with target text distribution having more impact than source text distribution. However, their experiments were limited to GPT-3 \citep{brown2020language} with only 100 test examples per direction. Our study substantially expands on this by extensively experimenting across the multiple axes mentioned earlier.

\noindent{\textbf{Perturbation and Robustness:}}
Text perturbation attacks \citep{xu-etal-2021-addressing,moradi-samwald-2021-evaluating,zhang-etal-2022-interpreting,niu-etal-2020-evaluating,salinas2024butterfly} are frequently used to assess the resilience of models to input alterations, distinguishing them from adversarial attacks \citep{michel-etal-2019-evaluation,Garg_2020,zhang-etal-2021-crafting} with their intent to simulate real-world noise rather than to explicitly deceive models. These attacks, applied at either word level \citep{zhengli2018iclr,cheng2020seq2sick} or character level \citep{belinkov2018synthetic,ebrahimi-etal-2018-adversarial,formento-etal-2023-using}, assess NMT model robustness through strategies like introducing random or linguistically aware perturbations, such as frequently confused characters. Additionally, perturbations may also involve random removal or addition of punctuations  \citep{formento-etal-2023-using}. Closest to our work is the work by \citet{moradi-samwald-2021-evaluating}, however, our approach differs from theirs as we only perturb demonstrations while leaving test examples as is.

%% file: content/methodology.tex
\section{Methodology}\label{sec:methodology}

Our experimentation first focuses on understanding whether in-context learning for MT is instruction-driven or example-driven, and then expands on the latter, exploring various aspects of examples. We categorize our experimentation along two principal axes: \textit{instruction perturbations} and \textit{demonstration perturbations}.

\subsection{Instruction Variations}\label{sec:instruction-perturb}

LLMs require extensive task descriptions for controlled generations and aligned outputs \citep{bsharat2023principled}. We intend to understand if instructions are really necessary or not and whether in-context exemplars can guide the model to perform the MT task in cases of suboptimal or confounding instructions. Keeping the demonstrations fixed, we experiment with the task instructions described below:

\begin{itemize}[label={},leftmargin=0pt]
    \item \textbf{Standard instruction:}
    The baseline condition, a default instruction, indicates the MT task is used.

    \item \textbf{No instruction:}
    Only in-context exemplars are used without any explicit instructions to see if the models can infer the task implicitly.
    
    \item \textbf{Generic instruction:}
    A non-task-specific generic instruction is used. This is similar to the previous setting, however, we add a generic instruction to infer the task from the exemplars.
    
    \item \textbf{Constrastive instruction:}
    An instruction with the opposite translation direction than the in-context exemplars is used to assess the model's susceptibility to off-target translations due to instruction changes.
    
    \item \textbf{Random instruction:}
    This includes a random instruction from a pool of non-translation tasks, while the in-context exemplars still reflect the translation task, to see if the model is misled into performing a different task.
\end{itemize}

\subsection{Demonstration Perturbations}\label{sec:homo-peturb}
In this set of experiments, we intend to investigate the impact of perturbing in-context demonstrations on the downstream MT performance, keeping instructions fixed. Specifically, our objective is to ascertain whether the model can effectively follow clear instructions and perform the tasks or whether the inclusion of suboptimal in-context exemplars adversely influences the subsequent downstream MT performance. To achieve this, we homogeneously either perturb the source or target distribution of in-context demonstrations by introducing different types of errors. Our perturbation methods are limited by a noise budget and influence the lexical (alterations to individual words), syntactic (alterations to structure or ordering of words), and semantic (alterations to meaning) properties of the in-context demonstrations and are listed below:

\begin{itemize}[label={},leftmargin=0pt]

    \item\textbf{Span Noise:}
    Random contiguous segments of the original text are modified by string operations such as deletion or replacement of characters within the selected spans similar to \citet{maurya2023utilizing} and \citet{joshi2020spanbert}.
    We consider the number of characters to be perturbed by uniformly selecting 1-3 gram spans and uniformly choosing to delete or replace with a single random character until the chosen budget is exhausted.
    
    \item\textbf{OCR:}
    Words from the original text are uniformly selected, and operations, such as fusion with adjacent words or splitting into two words to simulate noise, are performed, which might commonly be introduced in the pipelines involving OCR systems. 
    The noise percentage determines the degree of perturbation, affecting how many words are manipulated with a uniform probability of fusing consecutive words together or splitting a word into two parts.
    
    \item\textbf{Word Ordering:}
    The original word order of some portions of the original text is disrupted. 
    It is important to note that this perturbation can have different implications across different languages due to differing levels of word order flexibility. 
    We uniformly sample a set of words from the original text based on noise budget for reordering by generating a new order for these words.

    \item\textbf{Word Duplication:}
    Redundancy is added to the original text without altering its semantics by duplicating certain words. 
    We uniformly choose a set of words for duplication based on noise budget.
    
    \item\textbf{Punctuation:}
    The syntactic cues and rhythm of the original text are altered by either inserting or deleting existing punctuations present in the original text. 
    We uniformly add punctuation to words lacking it in addition and uniformly delete from the original text in case of deletion, given a noise budget.
\end{itemize}

\Cref{tab:perturbation_method_categorization} in \Cref{app:perturb-examples} provides a categorization of various perturbation methods considered as a part of this study with regard to the properties it impacts.

\subsection{Role of Demonstration Attributes}

While perturbations influence exemplar quality, there are other important attributes such as source distribution, target distribution, spatial proximity of the demonstration, and alignment between in-context demonstrations and test examples. We focus on these to further delve deeper into which attributes of an in-context demonstration play a critical role in downstream performance.

\begin{itemize}[label={},leftmargin=0pt]

\item\textbf{Heterogeneous Perturbations:} We perform heterogeneous perturbations to investigate if spatial proximity to the test example affects the downstream performance. Specifically, we consider cases involving a mix of clean and noisy demonstrations: $\{(k-1\ \mathrm{c}, 1\ \mathrm{n}), (1\ \mathrm{c}, k-1\ \mathrm{n}), (1\ \mathrm{n}, k-1\ \mathrm{c}), (k-1\ \mathrm{n}, 1\ \mathrm{c})\}$ where $c$ and $n$ indicate clean and noisy and $k$ denotes number of shots.

\item\textbf{Directionality:}
While the influence of test set directionality and translationese effects on traditional MT is well-explored \citep{zhang-toral-2019-effect,federmann-etal-2022-ntrex}, its impact on the performance of LLMs using in-context examples remains largely unexplored \citep{raunak-etal-2023-dissecting}. We explore this by choosing in-context exemplars from source-original and target-original sets to analyze whether leveraging target-original in-context examples is a superior choice than source-original in-context examples.

\item\textbf{Demonstrations from Allied Tasks as Proxy:}
When demonstrations for the specific task are not available during inference, a key question arises: \textit{Can demonstrations from a related task serve as a proxy for the model?} This is particularly relevant in MT, where obtaining demonstrations for every low-resource language pair might not be feasible. In this scenario, if the translation direction at test time is from language X to Y, we provide exemplars from language A to Y to see if these are sufficient to guide the model in generating language Y effectively. Furthermore, we also experiment with various auxiliary languages as the source for these in-context exemplars to determine if the choice of auxiliary language affects downstream performance. Specifically, we consider whether the auxiliary language a) matches the test time source language (baseline), b) matches the script of the test time source language and is not the test time source language, c) is English, d) is a randomly selected non-English language that differs in script from the test time source language.

\item\textbf{Misalignment:}
Effective in-context learners should extract pertinent information from the context and provide precise responses. However, this trait can be exploited by intentionally injecting misinformation into the examples. Therefore, we aim to understand if models are susceptible to such attacks. This differs from prior perturbation experiments (see \Cref{sec:homo-perturb}), where attacks targeted either source or target distribution without introducing misinformation through alignment alterations. Our experiment emulates a pivot translation scenario. We present two in-context examples: the first in language X with its English translation and the second demonstrating the translation of the English sentence into language Y. During testing, the model is tasked with translating from language X to Y, resembling a multi-hop reasoning process. Here, we consider 4 alignment types: a) all aligned, b) pivot misaligned, c) pivot and target misaligned, d) target misaligned. Types a) and b) are unperturbed cases, while c) and d) present perturbations or misinformation.

\end{itemize}

%% file: content/experiment_setup.tex
\section{Experimental setup}

This section describes the different pre-trained models, evaluation benchmarks, in-context prompts, and decoding hyperparameters used for our current study.

\begin{itemize}[label={},leftmargin=0pt]

\item \textbf{Models:} We conduct our experimentation on off-the-shelf open-source pre-trained LLMs like Llama 2 \citep{touvron2023llama2}, BLOOM \citep{bigscience-2022-bigscience} and their corresponding instruction-tuned variants, ALMA \citep{xu2023paradigm} and BLOOMZ \cite{muennighoff-etal-2023-crosslingual} respectively. Firstly, models like Llama 2 \cite{touvron2023llama2} are pre-dominantly trained with English or European languages and the tokenizer has poor fertility on Indic languages. Further, \citet{ahuja2023megaverse} show that Llama 2 70B \citep{touvron2023llama2} exhibits poor in-context MT performance on Indian languages. Hence, we evaluate Llama 2 \citep{touvron2023llama2} and ALMA \citep{xu2023paradigm} only on European languages and evaluate BLOOM \citep{bigscience-2022-bigscience} and BLOOMZ \cite{muennighoff-etal-2023-crosslingual} on the remaining three Indian languages. For consistency in our experiments, we add a task-specific fine-tuned baseline (BLOOM 7B FT) by fine-tuning the BLOOM 7B model on a subset of BPCC-seed data \citep{gala2023indictrans} using the same prompt structure as ALMA \citep{xu2023paradigm}. The fine-tuning hyperparameters are outlined in \Cref{app:fine-tune-bloom-7b-mt}. Due to computational constraints, we restrict our experimentation to 7B models for a fair comparison across the base and instruction-tuned variants. We adopt a greedy decoding strategy without sampling to generate a maximum of 256 tokens with early stopping to ensure reproducibility in our results.

\item \textbf{Languages:}
Broadly, our experimentation consists of 3 Indian languages like Bengali, Hindi, and Tamil, and 3 European languages like Czech, German, and Russian. However, we consider different sets of languages for certain experiments due to the nature of the experiment or the availability of the data. \Cref{tab:experiment-config} in \Cref{app:langs-considered} provides the languages considered, and the test set for each experiment.

\item \textbf{Benchmarks:}
FLORES-200 \citep{goyal-etal-2022-flores,nllbteam2022language} is a N-way parallel benchmark that covers 200 languages, including various low-resource languages. However, it is important to note that data from FLORES-200 \citep{goyal-etal-2022-flores,nllbteam2022language} has been consumed in the multitask fine-tuning xP3 mixture of BLOOMZ \citep{muennighoff-etal-2023-crosslingual}, suggesting data contamination. Therefore, we do not consider FLORES-200 for the evaluation of Indian languages and instead, we use the newly released IN22-Gen \citep{gala2023indictrans}. IN22-Gen is a 22-way parallel benchmark for Indian languages that focuses on demography-specific content and serves as an unbiased evaluation set for both BLOOM and BLOOMZ models.

\item \textbf{Prompt Details:}
In order to maintain consistency in terms of evaluation, we follow the same prompt that was used for fine-tuning ALMA \citep{xu2023paradigm} across all different models considered in this work (see \Cref{app:prompt-templates}). However, we find Llama 2 Chat models do not perform well with the above standard prompt in the monolithic format, which is similar to \citet{sclar2023quantifying}. Therefore, we resort to passing each demonstration as a user-assistant turn and task description in the system prompt in the chat format. We use uniformly sampled high-quality translation pairs from the FLORES-200 dev set to compose the in-context demonstrations with $k=\{1, 4, 8\}$ across various noise scales with $\delta=\{0, 0.1, 0.25, 0.5, 0.75\}$ for evaluating LLMs.

\item \textbf{Evaluation:}
Following \citet{sai-b-etal-2023-indicmt}, we use the ChrF++ metric \citep{popovic-2017-chrf}, computed using SacreBLEU\footnote{\texttt{sacreBLEU ChrF++ signature: \\nrefs:1|case:mixed|eff:no|tok:flores200|\\ smooth:exp|version:2.4.0}} \citep{post-2018-call}, as the primary metric for our analysis due to its strongest correlation with human judgments among automatic metrics.

\end{itemize}

%% file: content/results_and_analysis.tex
\section{Results and Analysis}\label{sec:results}

We now describe results that attempt to reveal the various factors that influence ICL for MT.

\begin{figure}[]
    \centering
    \includegraphics[width=\linewidth]{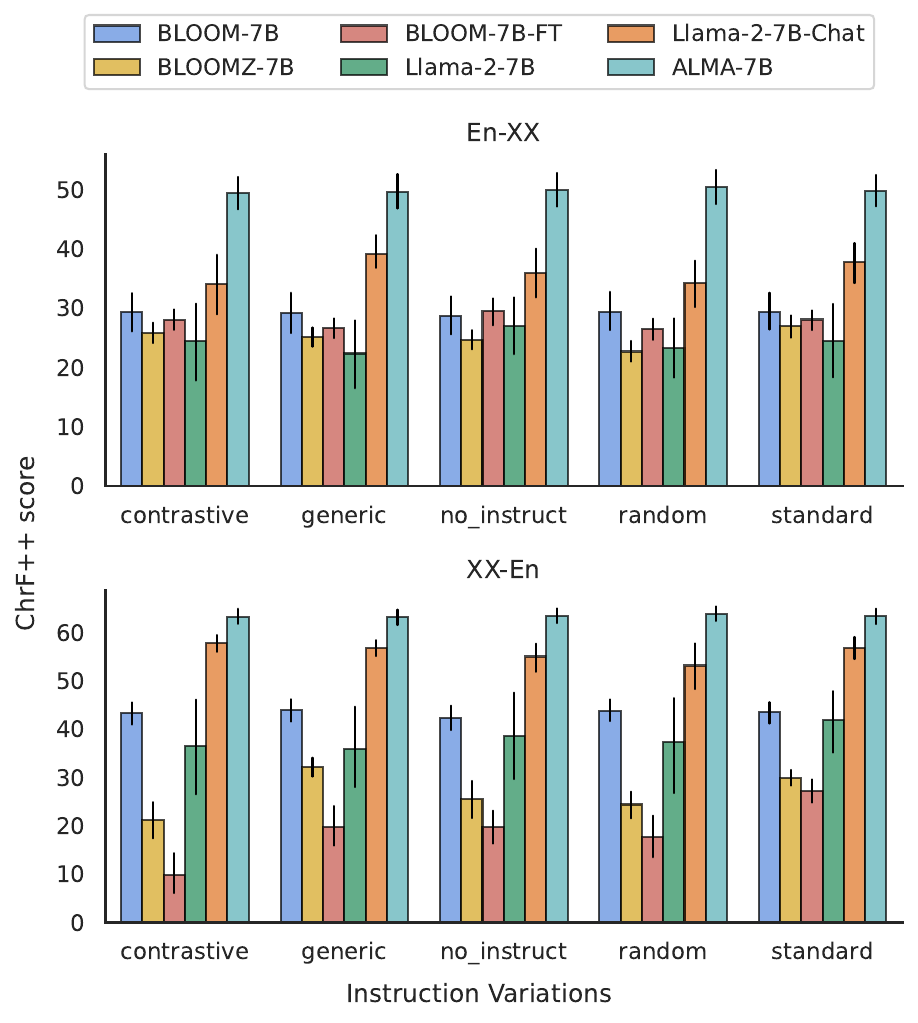}
    \caption{Comparison of ChrF++ scores for En-XX (top) and XX-En (bottom) across different instruction types for BLOOM and Llama 2 model families (averaged across different languages and shots).}
    \label{fig:instruct-trends}
\end{figure}

\subsection{Sensitivity to Instruction Variations}\label{sec:instruct-perturb}


\Cref{fig:instruct-trends} illustrates that base LLMs such as BLOOM-7B and Llama 2 7B exhibit less variability to instruction perturbations averaged across different shots. Furthermore, \Cref{fig:instruct-trends-shotwise} provides detailed results depicting performance variation across different shots subjected to instruction perturbation. We observe that chat models like BLOOMZ-7B and Llama 2 7B Chat show some sensitivity to instructions, particularly in a 1-shot setting, but this sensitivity diminishes as the number of shots increases. On the other hand, task-specific fine-tuned models like ALMA and BLOOM-7B-FT might be expected to be highly sensitive to instructions due to their training on data with specific instruction formats. Surprisingly, these models remain largely unaffected by perturbed instructions, even in a 1-shot setting. Although contrastive or random instructions aim to induce off-target or irrelevant generations, we observe that all the models consistently perform the translation task in the appropriate language, albeit with slightly reduced quality, highlighting that in-context examples primarily influence task recognition and learning.


\subsection{Example Perturbations}
Having established that models are mostly immune to instruction perturbations and are influenced more by examples, we focus on the latter.

\subsubsection{Homogeneous Perturbations}\label{sec:homo-perturb}

\begin{figure*}[ht]
    \centering
    \includegraphics[width=\textwidth]{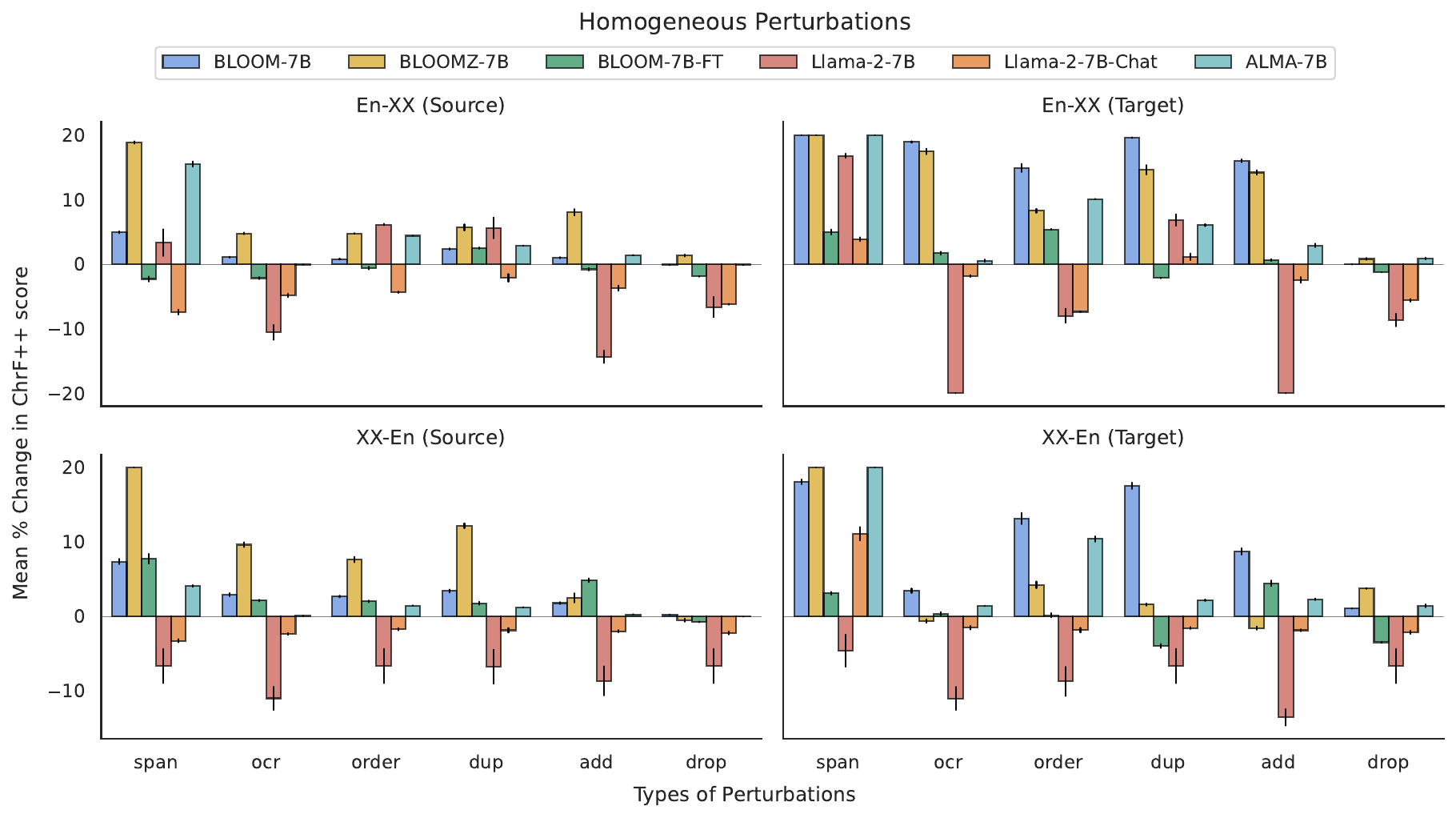}
    \caption{Mean percent change in ChrF++ score for each attack relative to the 0 noise baseline for each model across both translation directions (En-XX and XX-En) and both perturbation directions. Scores are averaged across shots and noise percentages ($\delta$). ``Span'' denotes span noise, ``order'' represents word order attack, ``dup'' signifies word duplication attack, ``add'' indicates a punctuation addition attack, and ``drop'' signifies punctuation removal attack. Positive values indicate the performance decreased post perturbation while the negative values indicate that performance increases post perturbation. Note: In certain cases, scores are bounded within minimum and maximum values for trend clarification.
    }
    \label{fig:homogeneous-trends}
\end{figure*}

\Cref{fig:homogeneous-trends} indicates that perturbing in-context examples appear to adversely impact performance despite clear instructions. It is important to note that these instructions do not suffice as substitutes for noisy in-context examples. Generally, perturbations to the target distribution have a more severe impact than those to the source distribution, however, we observe instances where the impact of the source distribution cannot be disregarded. For instance, the BLOOMZ-7B model and ALMA-7B are susceptible to source-side perturbations, necessitating dedicated experimentation to ascertain the significance of the source distribution. We explore this aspect further in \Cref{sec:allied}. Span noise emerges as the most detrimental perturbation across different model families, with the exception of BLOOM-7B-FT and Llama 2 7B models.

Across various perturbation attacks, we observe that the task-specific fine-tuned BLOOM-7B-FT model is most robust, followed by the BLOOM-7B model, while the multitask fine-tuned BLOOMZ-7B model is highly vulnerable. Llama 2 model generally demonstrates strong robustness to different attacks, except for the task-specific fine-tuned ALMA model, which is susceptible to span noise attack on the target side. Despite the relatively mild nature of word duplication attacks, the BLOOM-7B models surprisingly show high susceptibility. BLOOMZ-7B is robust to all attacks in XX-En translation, except for the span noise attack. Surprisingly, the Llama 2 7B model shows significant performance improvements when subjected to various attacks compared to the clean baseline across both translation and perturbation directions.

The current experiments primarily focus on homogeneous perturbations, indicating a detrimental impact on performance when in-context examples are perturbed uniformly. To investigate the potential influence of spatial proximity between perturbed in-context examples and the test examples, we delve into heterogeneous perturbations in \Cref{sec:hetero-perturb}. To limit the number of experiments due to computational constraints, we choose to exclude punctuation add and drop attacks, as the number of punctuations in a sentence is limited. 

\subsubsection{Heterogeneous Perturbations}\label{sec:hetero-perturb}

\begin{figure*}[!ht]
    \centering
    \includegraphics[width=\textwidth]{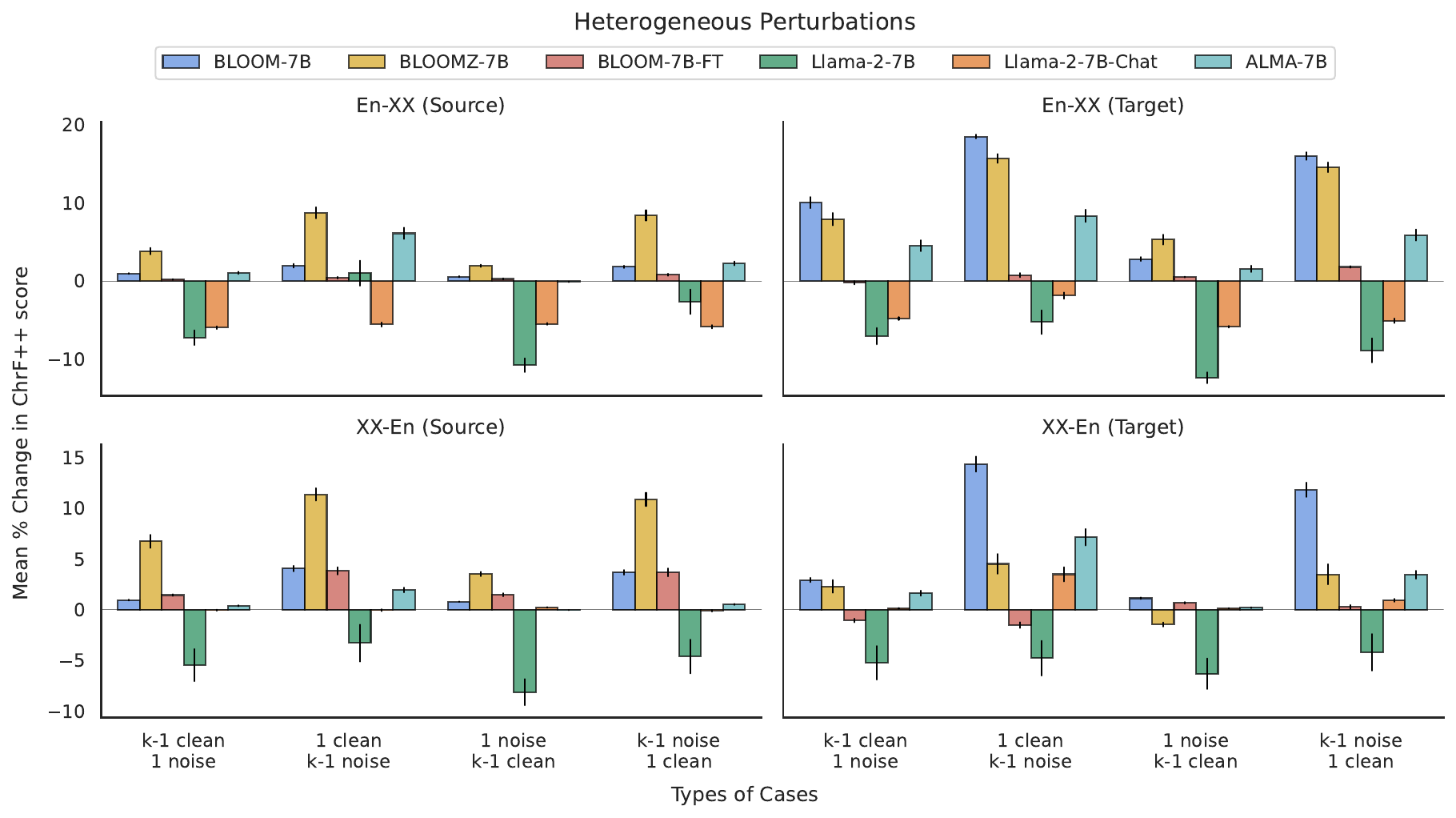}
    \caption{Mean percent change in ChrF++ score across different heterogeneous cases relative to the 0 noise baseline for each model across both translation directions (En-XX and XX-En) and both perturbation directions. Scores are averaged across attack types, shots and noise percentages ($\delta$). Positive values indicate the performance decreased post perturbation while the negative values indicate that performance increases post perturbation. Note: In certain cases, scores are bounded within minimum and maximum values for clarity in depicting overarching trends.}
    \label{fig:heterogeneous-trends}
\end{figure*}

We consider four cases categorized into two pairs based on the degree of noise perturbation: $\{(k-1\ \mathrm{c}, 1\ \mathrm{n}),(1\ \mathrm{n}, k-1\ \mathrm{c})$ and $\{(1\ \mathrm{n}, k-1\ \mathrm{c}), (k-1\ \mathrm{n}, 1\ \mathrm{c})\}$ where $c$ and $n$ indicates clean and noisy. We find that placing noisy examples closer to the test example has a more detrimental impact than placing them farther away (see \Cref{fig:heterogeneous-trends}). Among these cases, $\{(1\ \mathrm{c}, k-1\ \mathrm{n})\}$ is identified as the most detrimental. Consistent with prior findings, we note an improvement in Llama 2 7B performance when exposed to noise, with the most significant improvement occurring when the noisy examples are furthest from the test example (optimal conditions at $(k-1\ \mathrm{c}, 1\ \mathrm{n})$). Llama 2 7B Chat is fairly robust and demonstrates improvement upon perturbation, except for the most detrimental case. ALMA shows slight susceptibility across all variants, with the most severe impact observed in the most detrimental case. BLOOM-7B demonstrates robustness against source-side perturbations but is heavily affected by target-side perturbations, particularly when $k-1$ noisy examples are closer to the test example. BLOOMZ-7B is generally affected across all settings, with the most severe impact in the most detrimental case. Conversely, BLOOM-7B-FT also demonstrates a fair amount of robustness. Our experimental results provide empirical evidence to practitioners that in settings involving a mixed pool of noisy and clean in-context examples, the optimal approach is to position noisy examples at the beginning and cleaner examples closer to the test example to maximize performance.

\subsubsection{Demonstrations from Allied Tasks}\label{sec:allied}

\begin{figure}[t]
    \centering
    \includegraphics[width=\columnwidth]{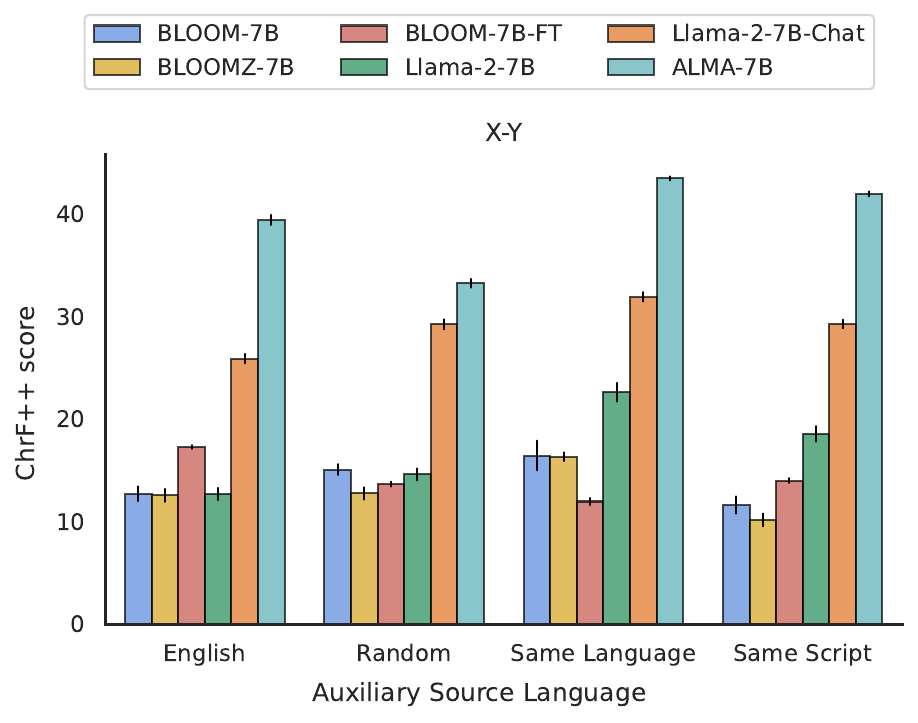}
    \caption{ChrF++ score of different models averaged across shots and translation directions comparing the choice of the auxiliary source language of the demonstration.}
    \label{fig:implicit-trends}
\end{figure}

Prior experiments (see \Cref{sec:instruct-perturb,sec:homo-perturb}) indicate that examples are more influential than instructions, and the target distribution of the example has broadly more importance than the source distribution. However, our focus has been limited to in-context examples of the same task during testing. To explore the limits of example-based learning, we aim to investigate if in-context examples from a different source language but the same target language suffice as a proxy for guiding the model in translating into the desired target language, thus probing the limits of example-based learning.

Our initial experiments, selecting an auxiliary language, showed that using in-context examples from a related task can generally serve as a suitable proxy for the target task, barring a few exceptions. To delve deeper, we considered 2 additional settings: using the same script as the test time source language and using English, the predominant language models were trained on, as the auxiliary source. \Cref{fig:implicit-trends} shows that the source distribution of in-context examples has a marginal effect on downstream MT performance. This suggests that any auxiliary source language can generally guide the model adequately, although there are a few exceptions, such as Llama 2 7B where using the same script as the auxiliary language outperforms English, and BLOOMZ-7B where a randomly chosen auxiliary language surpasses even using the same language. This suggests the limited impact of the choice of source language (source distribution). We conclude that in the absence of demonstrations from the exact pair as the test examples, demonstrations from any alternative source language can serve as a viable substitute, yielding comparable performance to that of the same source language baseline.

\subsection{Misalignment}\label{sec:misalignment}

\begin{figure}[t]
    \centering
    \includegraphics[width=\linewidth]{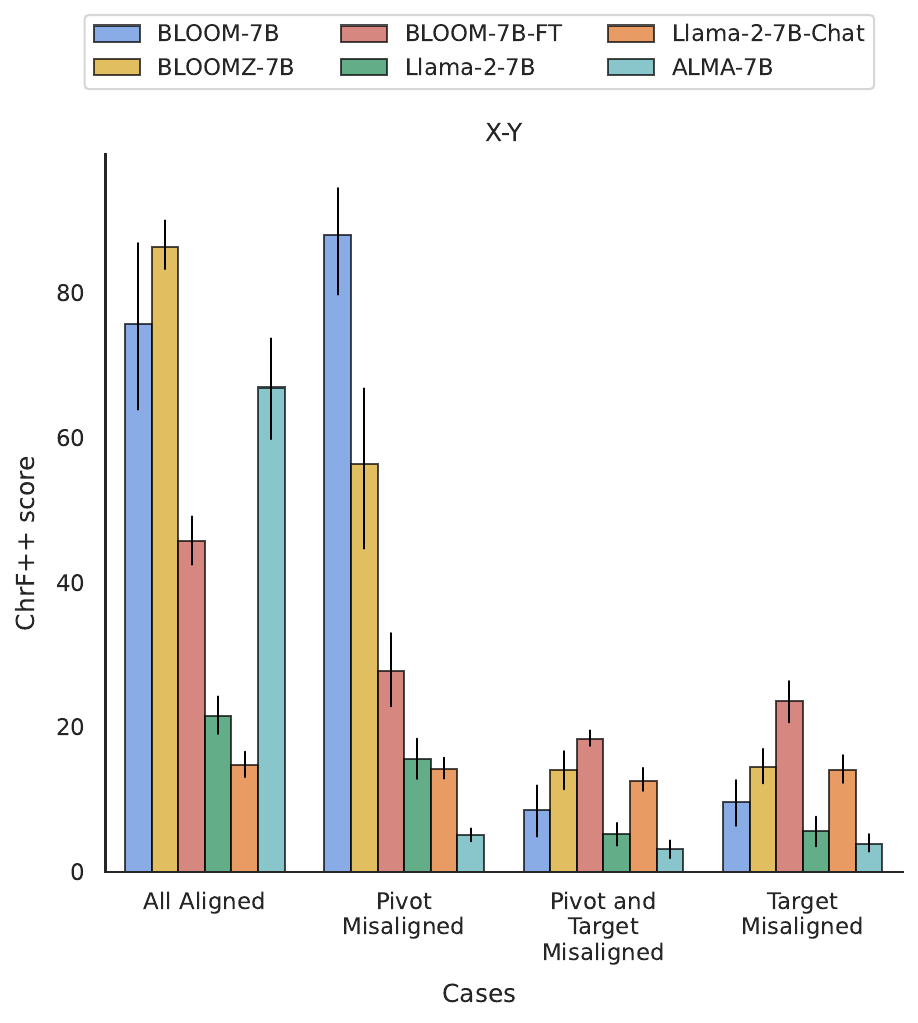}
    \caption{ChrF++ score of different models averaged translation directions comparing different forms of misalignment in a pivot-translation setting.}
    \label{fig:transitivity-trends}
\end{figure}

In this experimental setup, we aim to understand if semantic priors in models can protect against susceptibility to misinformation presented in context, rendering them effective and robust in-context learners. It is evident that most models, except Llama 2 7B Chat, can form transitive associations to varying degrees (see \Cref{fig:transitivity-trends}). Models from the BLOOM family and ALMA are particularly susceptible to extensive copy-pasting, suggesting they can be easily misled by in-context demonstrations, especially in cases of the target being misaligned. Notably, the ALMA model heavily relies on the English sentence as a pivot for these associations, and perturbing the English example also severely impacts performance despite the target translation being present in the context. Llama 2 7B exhibits poor translation quality, being unable to deduce answers from the context using transitive relations in aligned cases, and the performance further degrades in misaligned cases, indicating vulnerability to perturbations. In contrast, Llama 2 7B Chat remains largely unaffected by misinformation-based perturbations, showing robustness but poor downstream performance, indicating inadequacy in building transitive associations in the case of aligned examples. An ideal model should utilize its inherent biases to regulate its reliance on the information presented in context. Current 7B LLMs fall short in this regard, highlighting the necessity for future research to develop models that excel as in-context learners while minimizing susceptibility to misinformation within the context.

\subsection{Directionality of ICL Demonstrations}\label{sec:directionality}

\begin{figure}[]
    \centering
    \includegraphics[width=0.85\linewidth]{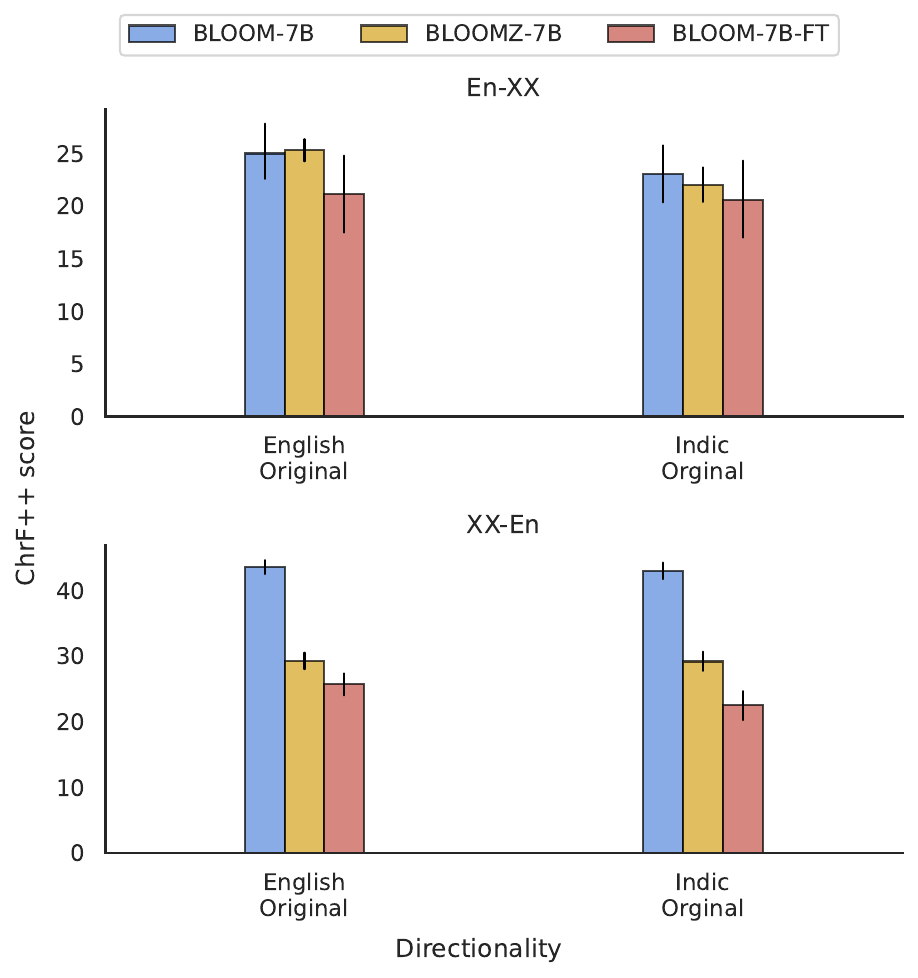}
    \caption{ChrF++ score of different models averaged across shots across translation directions comparing the choice of the source original and target original demonstrations.}
    \label{fig:directionality-trends}
\end{figure}

\Cref{sec:homo-perturb,sec:allied} shows the significance of target distribution; therefore, we further study if target-original demonstrations are superior to source-original demonstrations, depending on the translation direction. Focusing on Indic languages, we utilize an in-house multi-domain Indic Original set for sourcing Indic Original demonstrations and FLORES-200 dev set \citep{goyal-etal-2022-flores, nllbteam2022language} for English original ones. IN22-Gen \citep{gala2023indictrans}, which is English-original, serves as the test set. The Indic Original set\footnote{Will be released later as a held-out set of \url{https://www2.statmt.org/wmt24/multiindicmt-task.html}} serves as the target original set for En-XX direction and the source original set for XX-En, while FLORES-200 dev set \citep{goyal-etal-2022-flores, nllbteam2022language} acts as the target original set for XX-En direction and source original set for En-XX direction.

From \Cref{fig:directionality-trends}, across both En-XX and XX-En translation directions, we observe a minimal difference in the performance when demonstrations are chosen from the respective source original and target-original sets for all BLOOM family models considered. This suggests that the directionality of demonstrations may not have much significance on the downstream performance. However, it's important to acknowledge that our findings are limited to experiments conducted for Indic languages and using automatic metric-based evaluations, which may not sufficiently capture whether using target-original in-context examples has any impact on the fluency of the generation or not. Further investigation through human evaluation is necessary to ascertain this, an aspect we leave for future work.

%% file: content/conclusion.tex
\section{Conclusion}

Although ICL has demonstrated good performance and is widely employed across different tasks, there is limited understanding of the aspects that influence the downstream performance of ICL. In this work, we aim to bridge this gap for MT by investigating the central question of whether ICL for MT is example-driven or instruction-driven and subsequently examining the role of different aspects of ICL. We find that: (1) ICL is primarily example-driven and choice of instruction has limited impact, with target distribution of the demonstrations being the most influential (2) Perturbation to demonstrations is generally detrimental but can have a regularization effect in some cases (3) Spatial proximity to the test example is an important factor in ICL (4) Demonstrations having the same target language can serve as a reliable proxy and choice of the source language of the demonstration is inconsequential 
(5) The directionality of the demonstrations has minimal impact.
(6) ICL can potentially be exploited to mislead even smaller models, highlighting the need for further research to make models robust in-context learners.

%% file: content/limitations.tex
\section{Limitations}
Our perturbation experiments focused on only 5 non-linguistic perturbations, which involved random textual attacks that were mostly uniform and language-agnostic in nature. However, it would also be interesting to explore linguistically aware perturbations such as causality alternation, entity replacement, and number replacement, similar to \citet{chen-etal-2023-xsim}. This would test if the model is susceptible to more fine-grained and subtle errors. Additionally, our investigation was limited to the 7B model scale due to computational constraints, and exploring whether our current findings generalize across larger model scales would provide us a more comprehensive understanding of whether ICL capabilities models vary with scale. Furthermore, while the focus of this study was on the MT task, subsequent research should examine the transferability of these insights to diverse natural language generation tasks.

%% file: content/ethics.tex
\section{Ethical Considerations}

\noindent\textbf{Potential toxic and hateful outputs:} Our work focuses on the robustness of models to various factors influencing MT via ICL. One aspect of our exploration is via perturbation which may lead to hallucinations as well as generate toxic and hateful content as a result of the model's distributions being perturbed. This may even be applicable to other generative tasks but that is not the intention of our work.

\noindent\textbf{Safety Circumvention and Jailbreaking:} Adversarial perturbations might be able to circumvent a model's safety parameters and enable the generation of biased and harmful outputs. We plan to release our perturbation scripts and resultant perturbed data for research, and we do not intend it to be used to perform adversarial attacks in practice intended to undo the security features, also known as jailbreaking, of a model.

%% file: content/acknowledgements.tex
\section{Acknowledgements}
We sincerely thank Varun Gumma (SCAI Fellow, Microsoft Research) for the insightful discussions and his generous review of this draft, which helped us organize our ideas more effectively and improve its overall readability.

%% file: appendix/A.tex
\section{Fine-tuning BLOOM 7B on MT}\label{app:fine-tune-bloom-7b-mt}

We fine-tune BLOOM \citep{bigscience-2022-bigscience} 7B model on a human-annotated subset of BPCC \citep{gala2023indictrans}, namely BPCC-seed and this forms the task-specific fine-tuned baseline for the BLOOM family. \Cref{tab:bloom-7b-ft-data-stats} and \Cref{tab:bloom-7b-ft-hyperparams} list the data scales and hyperparameters used for fine-tuning respectively.

\begin{table}[h]
\centering
\begin{tabular}{lr}
\toprule
Language & Bitext pairs  \\
\midrule
Bengali & 50K \\
Hindi   & 50K \\
Marathi & 50K \\
Tamil   & 30.6K \\
Telugu  & 36.9K \\  
\bottomrule
\end{tabular}
\caption{Statistics of the sample of the English-centric BPCC-Seed data \citep{gala2023indictrans} used for fine-tuning the BLOOM 7B model. Note that the data was formatted in both directions (En-XX and XX-En) directions and used for joint fine-tuning.}
\label{tab:bloom-7b-ft-data-stats}
\end{table}

\begin{table}[h]
\centering
\begin{tabular}{lr}
\toprule
Hyperparameters & Value  \\
\midrule
Batch size & $128$ \\
Learning rate & $5e-5$ \\
Epochs & $4$ \\
Scheduler & linear \\
Warmup ratio & $0.03$ \\
Weight decay & $0.$ \\
\bottomrule
\end{tabular}
\caption{Hyperparameters for fine-tuning the BLOOM 7B model on MT task.}
\label{tab:bloom-7b-ft-hyperparams}
\end{table}

\section{Use of model-based metrics}
\label{sec:comet}
\begin{figure}[]
    \centering
    \includegraphics[width=\linewidth]{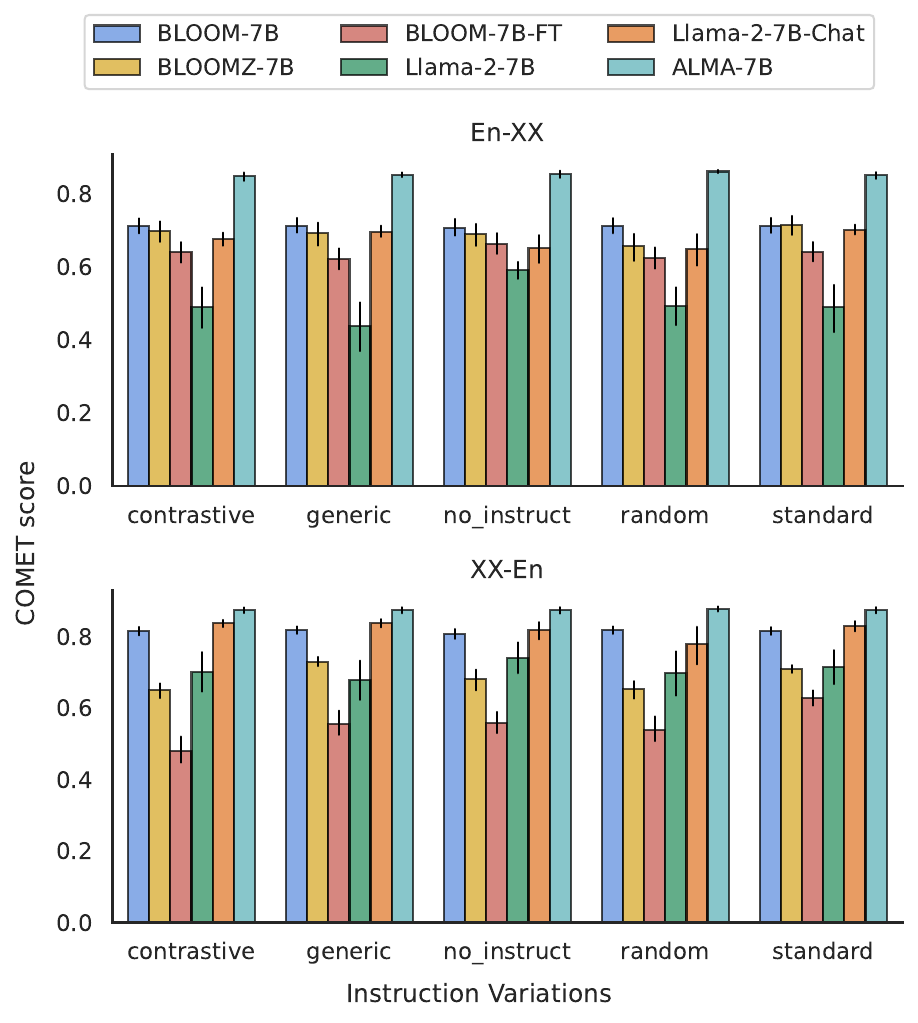}
       \caption{Comparison of COMET scores for En-XX (top) and XX-En (bottom) across different instruction types for BLOOM and Llama 2 model families (averaged across different languages and shots).}
    \label{fig:instruct-trends-comet}
\end{figure}

Model-based metrics such as COMET \citep{rei-etal-2022-comet} have demonstrated a good correlation with human judgments across several languages \citep{kocmi-etal-2021-ship}. However, these metrics may not be well-calibrated for all the languages considered in this study \citep{gala2023indictrans}. Furthermore, the scale of our experiments makes it impractical to compute model-based metrics like COMET due to the additional computational overhead needed.
Consequently, we use the best string-based metric, ChrF++ \citep{popovic-2017-chrf} as outlined in \citep{sai-b-etal-2023-indicmt}. For completeness, we also compute the COMET scores using the \texttt{wmt22-comet-da}\footnote{\url{https://huggingface.co/Unbabel/wmt22-comet-da}} model for one of our primary experimental setups (instruction perturbation) and found the trends to be consistent with the ChrF++ metrics. This suggests that ChrF++ can serve as a reliable proxy, indicative of the overall trends. \Cref{fig:instruct-trends-comet} shows the trends in COMET scores for the instruction perturbation experiment, clearly mirroring the trends in ChrF++ scores presented in \Cref{fig:instruct-trends}.

\section{Prompt Templates}\label{app:prompt-templates}

We use the standard prompt template (\Cref{fig:mt_prompt_template}) for most of our experiments except for experiments related to \Cref{sec:allied,sec:misalignment}. In the case of experiments in \Cref{sec:allied}, we do not specify the source language in the instruction similar to the prompt illustrated in \Cref{fig:allied_prompt_template}.
For the experiments in \Cref{sec:misalignment}, we use the prompt template specified in \Cref{fig:misalignment_prompt_template}.

\begin{figure}[h]
\centering
\begin{promptbox}
Translate this from \{src\_lang\} into \{tgt\_lang\}: \\ \\ 
\{src\_lang\}: \{src\_text\} \\
\{tgt\_lang\}: \{tgt\_text\} \\ \\ 
$\cdots$ \\
$\cdots$ \\

\{src\_lang\}: \{src\_text\} \\
\{tgt\_lang\}: 
\end{promptbox}
\caption{Standard prompt for translation}
\label{fig:mt_prompt_template}
\end{figure}

\begin{figure}[h]
\centering
\begin{promptbox}
Translate this into \{tgt\_lang\}: \\ \\ 
\{aux\_lang\}: \{aux\_text\} \\
\{tgt\_lang\}: \{tgt\_text\} \\ \\ 

$\cdots$ \\
$\cdots$ \\

\{src\_lang\}: \{src\_text\} \\
\{tgt\_lang\}: 
\end{promptbox}
\caption{Prompt for Allied Task Setup}
\label{fig:allied_prompt_template}
\end{figure}

\begin{figure}[!h]
\centering
\begin{promptbox}
Translate this from \{src\_lang\} into \{pivot\_lang\}: \\ \\ 
\{src\_lang\}: \{src\_text\} \\
\{pivot\_lang\}: \{pivot\_text\} \\ \\ 

Translate this from \{pivot\_lang\} into \{tgt\_lang\}: \\ \\ 
\{pivot\_lang\}: \{pivot\_text\} \\
\{tgt\_lang\}: \{tgt\_text\} \\ \\ 

Translate this from \{src\_lang\} into \{tgt\_lang\}: \\ \\ 
\{src\_lang\}: \{src\_text\} \\
\{tgt\_lang\}: 
\end{promptbox}
\caption{Misalignment prompt for translation}
\label{fig:misalignment_prompt_template}
\end{figure}

\section{Instruction Templates}
We outline the different types of instructions considered in the instruction perturbation experiments mentioned in \cref{sec:instruct-perturb}. For each type of instruction, an example is provided in \Cref{app:prompt-templates}. Additionally, for the random instruction type, any one of the prefix instructions is selected at random.

\begin{figure*}[h]
\centering
\begin{promptbox}
\# Standard \\ \\
{\color{blue!70} Translate this from \{src\_lang\} into \{tgt\_lang\}:} \\ \\

\# Generic \\ \\
{\color{red!70} Perform the task based on the examples provided:} \\ \\

\# Random \\ \\
{\color{purple!70} Complete the description with an appropriate ending: \\ \\
I am hesitating between two options. Help me choose the more likely cause or effect. \\ \\
Generate a headline for the following article(s) as accurately as possible. \\ \\
Predict the sentiment of the review. The possible choices for the sentiment are: 'positive' and 'negative'. \\ \\
Answer whether the hypothesis is more likely to be true (entailment), false (contradiction), or unknown (neutral) based on the given premise. \\ \\
The following are multiple choice questions (with answers) about subjects. \\
} \\ \\

\# Contrastive \\ \\
{\color{brown!70} Translate this from \{tgt\_lang\} into \{src\_lang\}:} \\
\end{promptbox}
\caption{Different types of instructions}
\label{fig:instruction_templates}
\end{figure*}

\section{Examples of Perturbation Variants}\label{app:perturb-examples}

\Cref{tab:perturbation_method_categorization} categorizes the different perturbations used in this study based on attributes they impact, along with an example.

\label{subsec:example_perturb_attack}
\begin{table*}[!t]
\small
\centering
\begin{tabular}{lcccl}
\toprule
Perturbation Method   & Lexical & Syntactic & Semantic & Example \\
\midrule
\multirow{2}{*}{Clean}   &   &   &   &   Wow! That place is so wonderful, and I would    \\
                        &   &   &   &   love to go there again.    \\
\midrule

\multirow{2}{*}{Span Noise} &    \multirow{2}{*}{$\checkmark$}     &   \multirow{2}{*}{$\checkmark$}     &   \multirow{2}{*}{$\times$}  &   Wow! That place is po wonde9l, a I uld \\
                            &   &   &   &   love to go thyre again.    \\
\midrule

\multirow{2}{*}{OCR}    &    \multirow{2}{*}{$\checkmark$}     &   \multirow{2}{*}{$\times$}    &   \multirow{2}{*}{$\times$} &   Wow!That place isso wonderful, and I would  \\
                        &   &   &   &  lo ve to go there again .  \\
\midrule

\multirow{2}{*}{Word Ordering}  &    \multirow{2}{*}{$\times$}     &   \multirow{2}{*}{$\checkmark$}    &   \multirow{2}{*}{$\checkmark$}     &   Wow! and place is so to would I That wonderful, \\
                                &   &   &   &     love go there again.    \\
\midrule

\multirow{2}{*}{Word Duplication} &    \multirow{2}{*}{$\checkmark$}    &   \multirow{2}{*}{$\times$}      &   \multirow{2}{*}{$\checkmark$}    &  Wow! That place is is so so wonderful, and I I would   \\
                                &   &   &   &  love to go there again. again.   \\
\midrule

\multirow{2}{*}{Punctuation$_{add}$}  &    \multirow{2}{*}{$\checkmark$}    &   \multirow{2}{*}{$\checkmark$}      &   \multirow{2}{*}{$\checkmark$}   &   Wow! That\% place is so wonderful, and I would" \\
                                    &   &   &   &  love. to go there again.   \\
\midrule

\multirow{2}{*}{Punctuation$_{drop}$} &    \multirow{2}{*}{$\checkmark$}    &   \multirow{2}{*}{$\checkmark$}     &   \multirow{2}{*}{$\checkmark$}   &   Wow That place is so wonderful, and I would   \\
                                      &   &   &   &  love to go there again    \\
\bottomrule
\end{tabular}
\caption{Categorization of different perturbation methods for the different attributes.}
\label{tab:perturbation_method_categorization}
\end{table*}

\section{Languages and Directions Considered}\label{app:langs-considered}

\Cref{tab:experiment-config} describes the languages considered and respective benchmarks used for each experiment conducted as a part of this study.

\begin{table*}[]
\small
\centering
\begin{tabular}{lccccl}
\toprule
\textbf{Experiment} &
  \textbf{Models} &
  \textbf{\begin{tabular}[c]{@{}c@{}}Translation\\ direction\end{tabular}} &
  \textbf{Test set} &
  \textbf{In-context set} &
  \textbf{Languages} \\
\midrule
\multirow{2}{*}{\begin{tabular}[c]{@{}c@{}}Instruction\\variation\end{tabular}} &
  Llama 2 &
  \multirow{2}{*}{En-X} &
  Flores200 &
  Flores200 &
  ces\_Latn, deu\_Latn, rus\_Cyrl \\
 &
  BLOOM &
   &
  IN22-Gen &
  Flores200 &
  ben\_Beng, hin\_Deva, tam\_Taml \\
\midrule
\multirow{2}{*}{\begin{tabular}[c]{@{}c@{}}Demonstration\\perturbation\end{tabular}} &
  Llama 2 &
  \multirow{2}{*}{En-X} &
  Flores200 &
  Flores200 &
  ces\_Latn, deu\_Latn, rus\_Cyrl \\
 &
  BLOOM &
   &
  IN22-Gen &
  Flores200 &
  ben\_Beng, hin\_Deva, tam\_Taml \\
\midrule
\multirow{2}{*}{Directionality} &
  BLOOM &
  \multirow{2}{*}{En-X} &
  IN22-Gen &
  \begin{tabular}[c]{@{}c@{}}Flores200\\ IndicOG set\end{tabular} &
  ben\_Beng, guj\_Gujr, hin\_Deva, tel\_Telu \\
 &
  BLOOM &
   &
  IN22-Gen &
  \begin{tabular}[c]{@{}c@{}}Flores200\\ IndicOG set\end{tabular} &
  ben\_Beng, guj\_Gujr, hin\_Deva, tel\_Telu \\
\midrule
\multirow{2}{*}{\begin{tabular}[c]{@{}c@{}}Demonstrations from\\ allied task as proxy\end{tabular}} &
  Llama 2 &
  \multirow{2}{*}{\begin{tabular}[c]{@{}c@{}}En-X\\ X-Y\end{tabular}} &
  Flores200 &
  Flores200 &
  \begin{tabular}[c]{@{}c@{}}ces\_Latn - rus\_Cyrl \\ (deu\_Latn, eng\_Latn, hin\_Deva)\\ \\ deu\_Latn - rus\_Cyrl \\ (ces\_Latn, eng\_Latn, hin\_Deva)\\ \\ srp\_Cyrl - deu\_Latn \\ (rus\_Cyrl, eng\_Latn, hin\_Deva)\\ \\ srp\_Cyrl - ces\_Latn\\ (rus\_Cyrl, eng\_Latn, hin\_Deva)\end{tabular} \\
 &
  BLOOM &
   &
  IN22-Gen &
  Flores200 &
  \begin{tabular}[c]{@{}c@{}}mar\_Deva - tam\_Taml\\ (hin\_Deva, eng\_Latn, ben\_Beng)\\ \\ asm\_Beng - hin\_Deva\\ (ben\_Beng, eng\_Latn, tam\_Taml)\end{tabular} \\
\midrule
\multirow{2}{*}{Transitivity} &
  Llama 2 &
  \multirow{2}{*}{X-Y} &
  Flores200 &
  Flores200 &
  ces\_Latn, deu\_Latn, rus\_Cyrl \\
 &
  BLOOM &
   &
  IN22-Gen &
  Flores200 &
  ben\_Beng, hin\_Deva, tam\_Taml \\
\bottomrule
\end{tabular}
\caption{Details about the models, test sets, in-context sets and languages considered for different experiments. LLama2 family indicates Llama2-7B, Llama2-Chat-7B, ALMA while BLOOM family indicates BLOOM-7B, BLOOMZ-7B and a task-specific fine-tuned BLOOM model on MT. En-X in the translation direction indicates English-centric evaluation, while X-Y indicates non-English-centric evaluation. FLORES200 in the test set column indicates the FLORES200 devtest set, while in the in-context set column indicates FLORES200 dev set.}
\label{tab:experiment-config}
\end{table*}

\section{Additional results}
\begin{figure*}[]
    \centering
    \includegraphics[width=\textwidth]{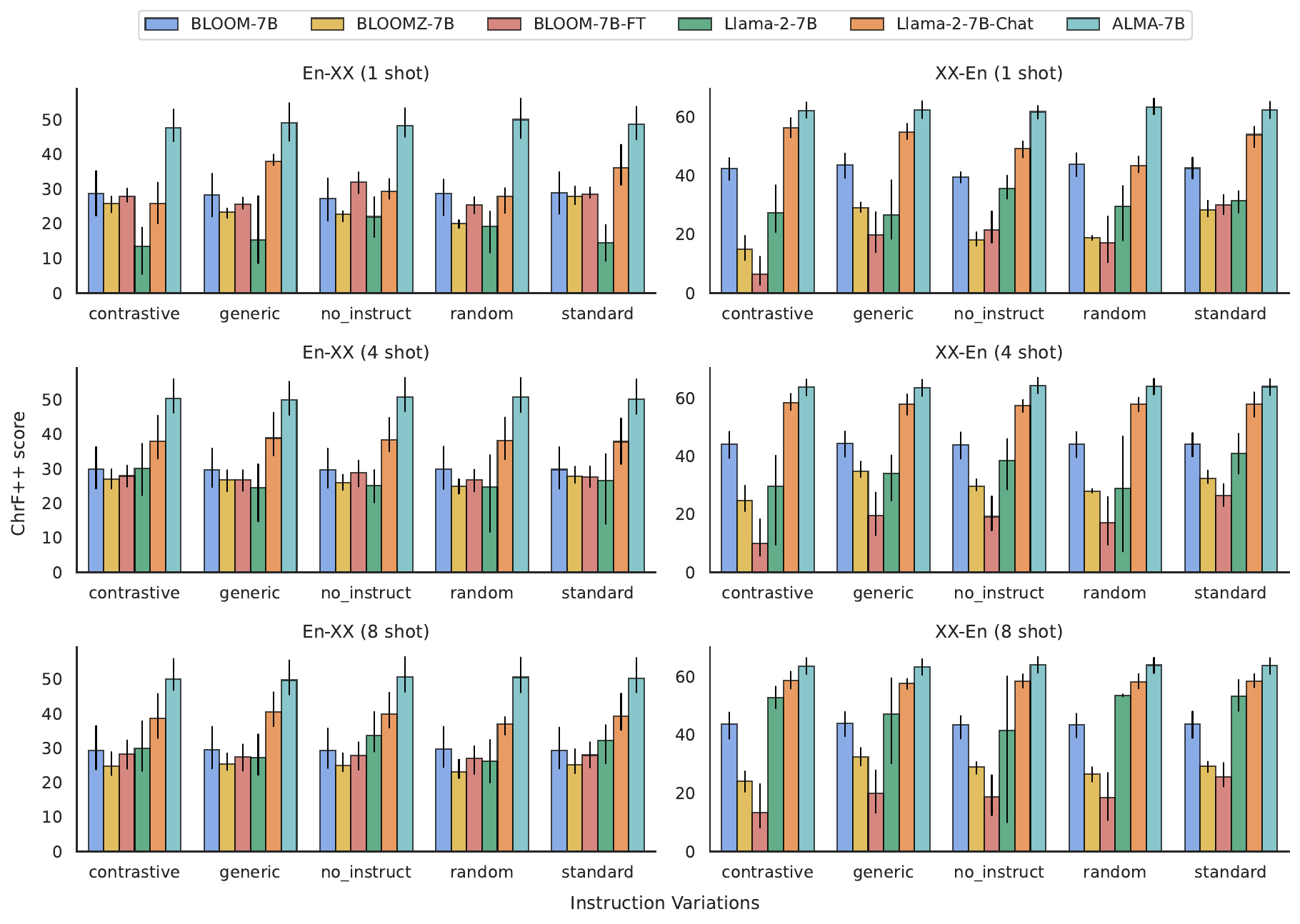}
    \caption{Shot-wise comparison of ChrF++ scores for En-XX (left) and XX-En (right) across different instruction types for BLOOM and Llama 2 model families (averaged across different languages).}
    \label{fig:instruct-trends-shotwise}
\end{figure*}

\begin{figure*}[]
    \centering
    \includegraphics[width=\textwidth]{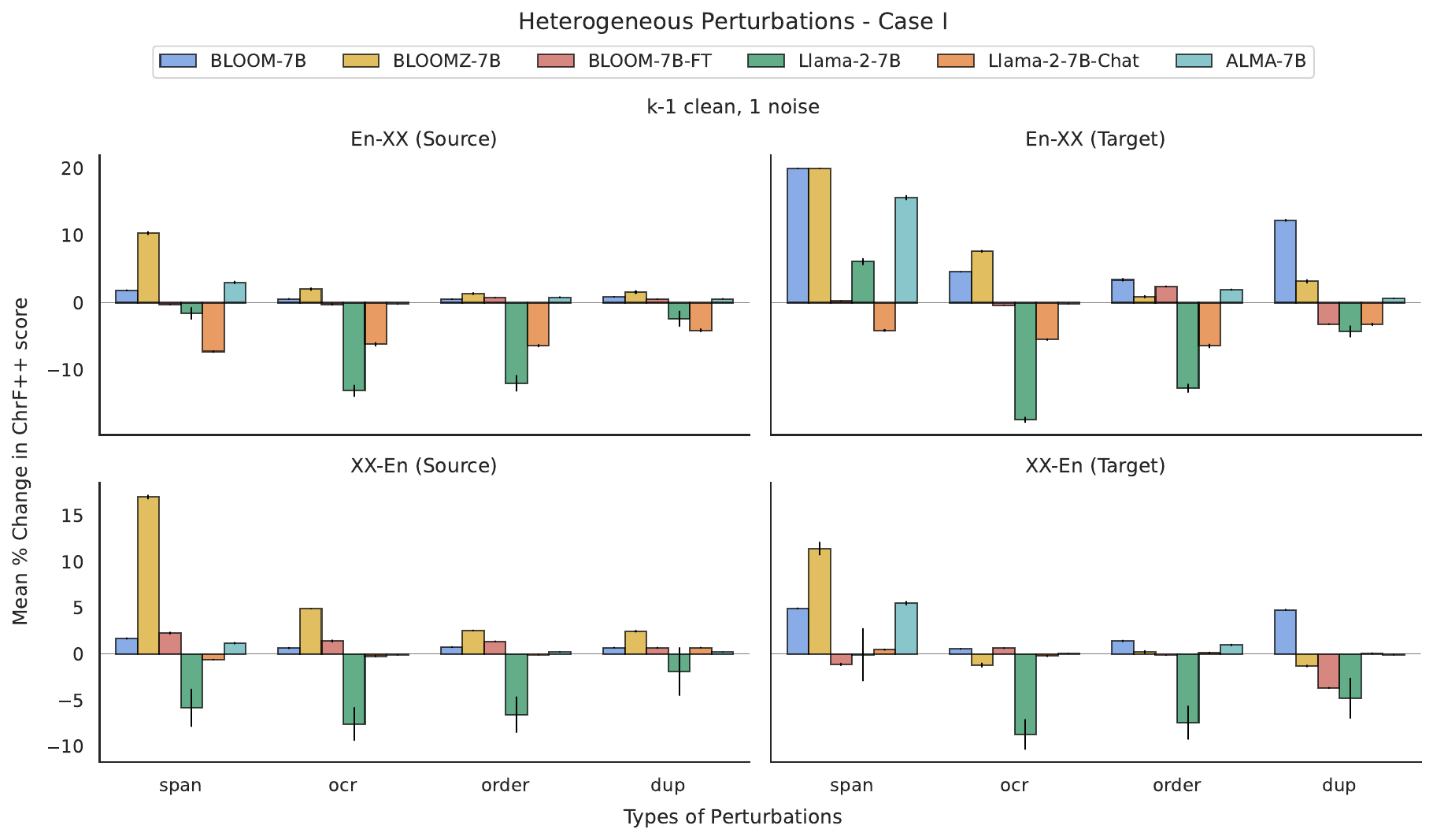}
    \caption{Mean percent change in ChrF++ score across heterogeneous case I relative to the 0 noise baseline for each model across both translation directions (En-XX and XX-En) and both perturbation directions. Scores are averaged across attack types, shots, and noise percentages. Positive values indicate the performance decreased post perturbation while the negative values indicate that performance increases post perturbation. Note: In certain cases, scores are bounded within minimum and maximum values for clarity in depicting overarching trends.}
    \label{fig:heterogeneous-I-trends}
\end{figure*}

\begin{figure*}[]
    \centering
    \includegraphics[width=\textwidth]{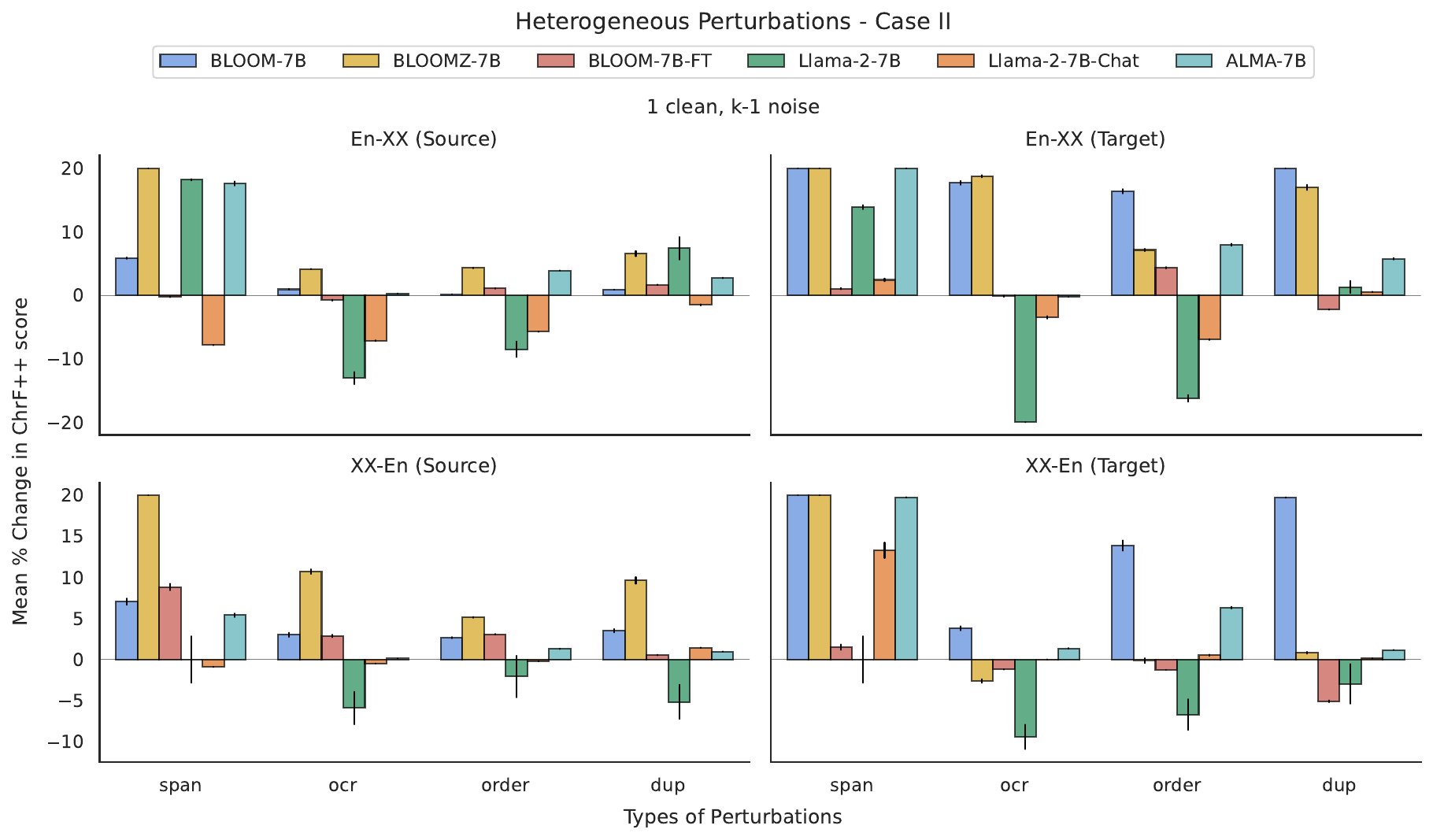}
    \caption{Mean percent change in ChrF++ score across heterogeneous case II relative to the 0 noise baseline for each model across both translation directions (En-XX and XX-En) and both perturbation directions. Scores are averaged across attack types, shots, and noise percentages. Positive values indicate the performance decreased post perturbation while the negative values indicate that performance increases post perturbation. Note: In certain cases, scores are bounded within minimum and maximum values for clarity in depicting overarching trends.}
    \label{fig:heterogeneous-II-trends}
\end{figure*}

\begin{figure*}[]
    \centering
    \includegraphics[width=\textwidth]{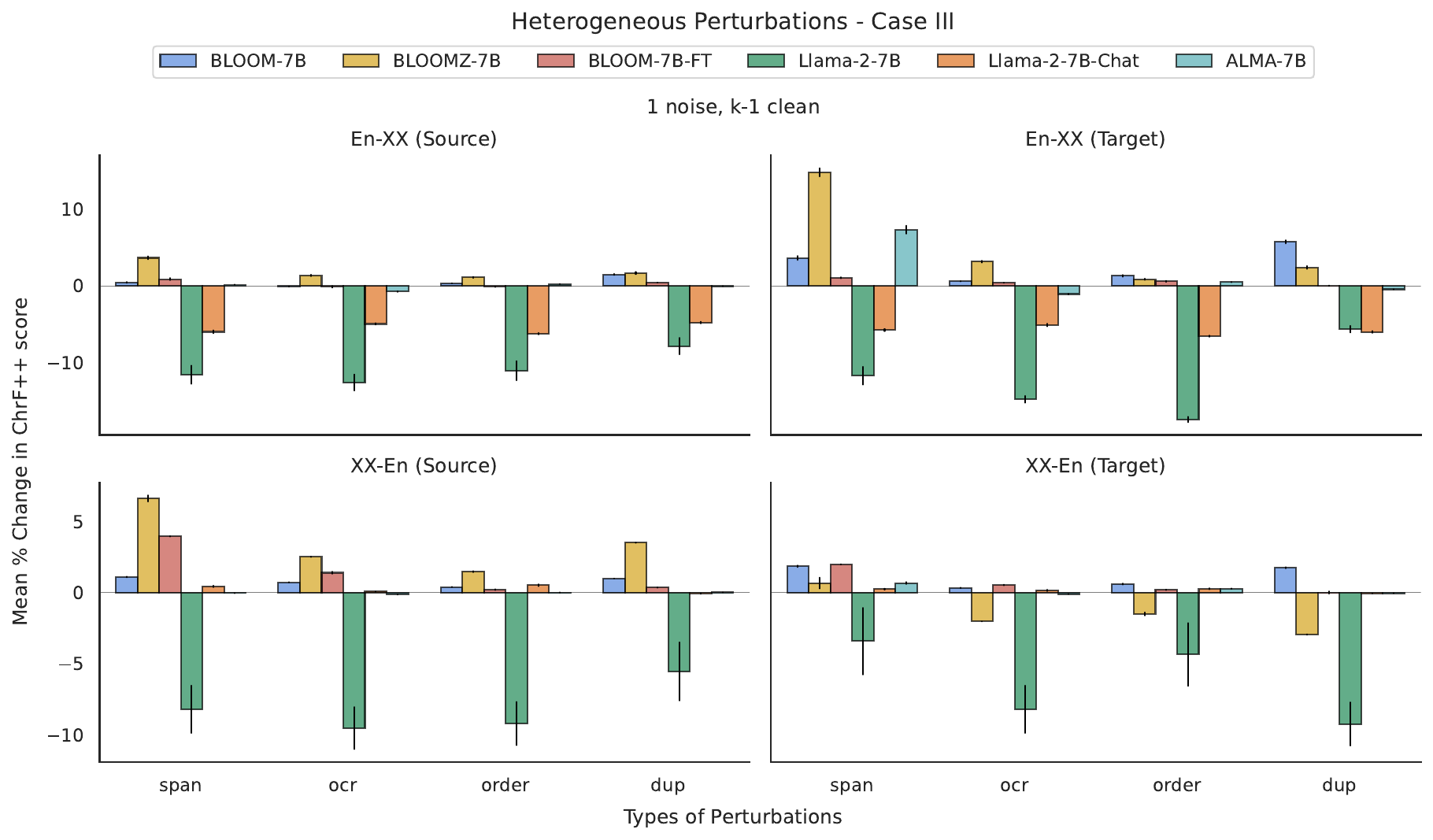}
    \caption{Mean percent change in ChrF++ score across heterogeneous case III relative to the 0 noise baseline for each model across both translation directions (En-XX and XX-En) and both perturbation directions. Scores are averaged across attack types, shots, and noise percentages. Positive values indicate the performance decreased post perturbation while the negative values indicate that performance increases post perturbation. Note: In certain cases, scores are bounded within minimum and maximum values for clarity in depicting overarching trends.}
    \label{fig:heterogeneous-III-trends}
\end{figure*}

\begin{figure*}[]
    \centering
    \includegraphics[width=\textwidth]{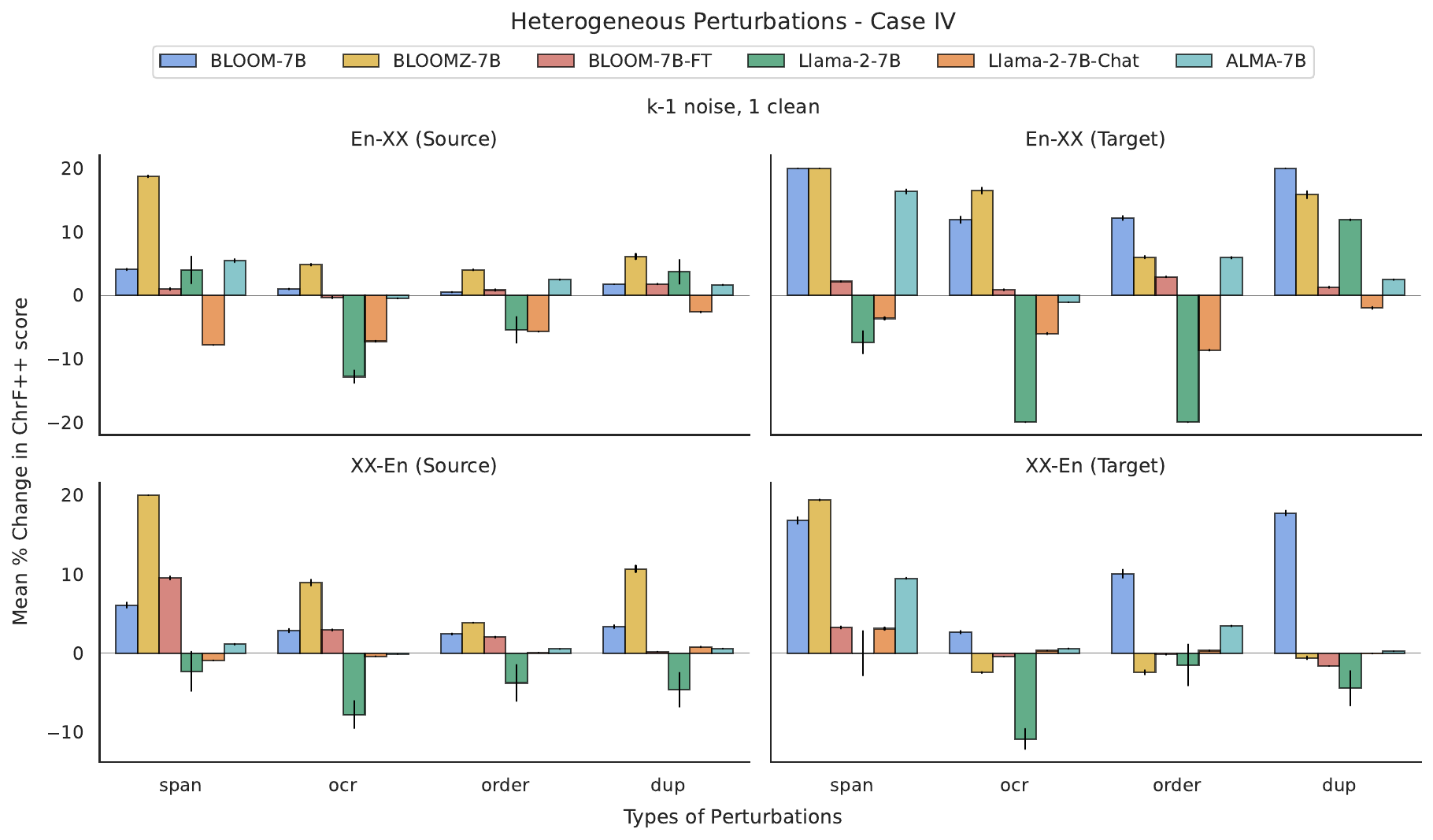}
    \caption{Mean percent change in ChrF++ score across heterogeneous case IV relative to the 0 noise baseline for each model across both translation directions (En-XX and XX-En) and both perturbation directions. Scores are averaged across attack types, shots, and noise percentages. Positive values indicate the performance decreased post perturbation while the negative values indicate that performance increases post perturbation. Note: In certain cases, scores are bounded within minimum and maximum values for clarity in depicting overarching trends.}
    \label{fig:heterogeneous-IV-trends}
\end{figure*}

\begin{figure*}[]
    \centering
    \includegraphics[width=0.7\textwidth]{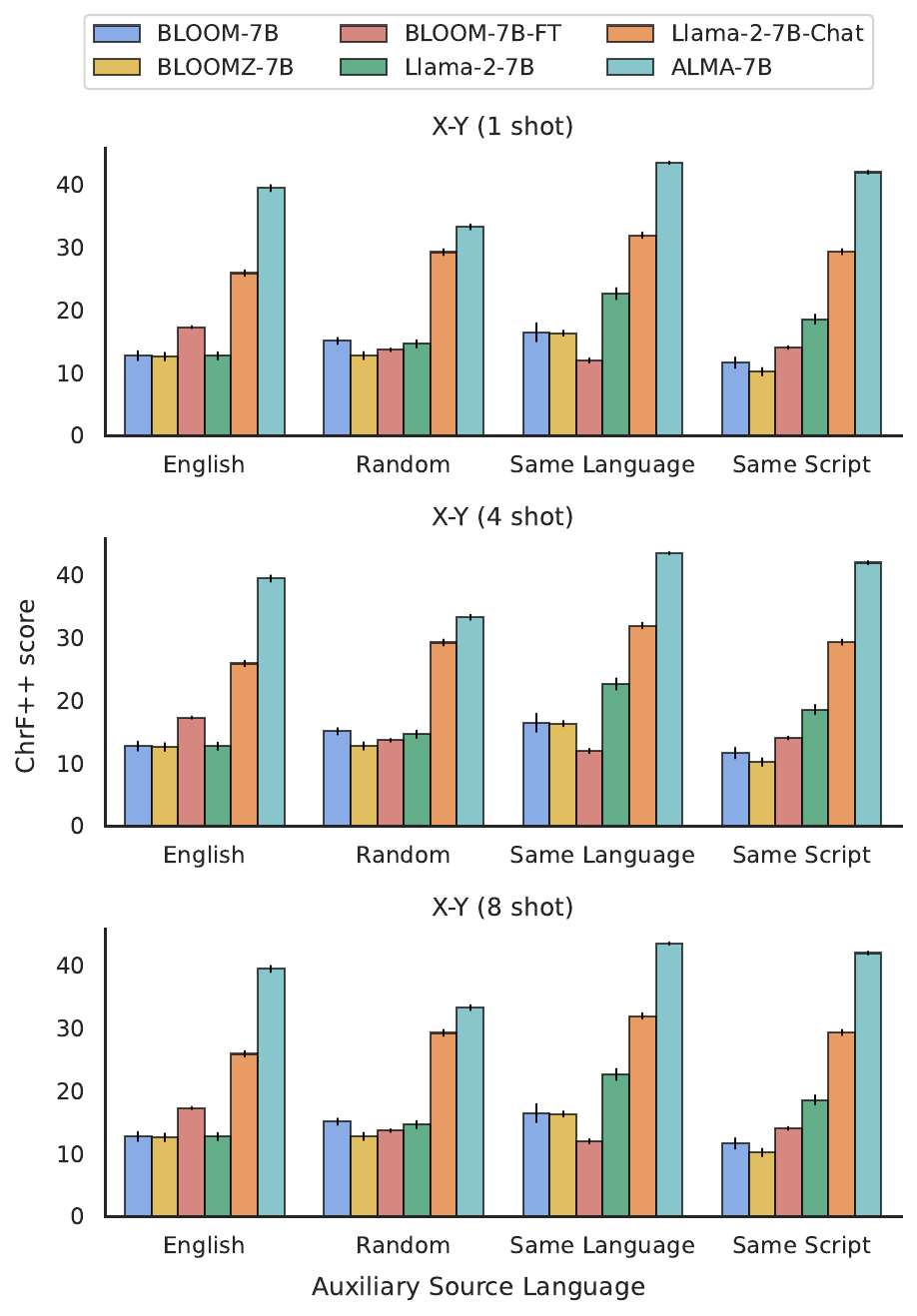}
    \caption{Shot-wise comparison of ChrF++ score of different models averaged across translation directions comparing the choice of the auxiliary source language of demonstrations.}
    \label{fig:implicit-trends-shotwise}
\end{figure*}

\begin{figure*}[]
    \centering
    \includegraphics[width=0.7\textwidth]{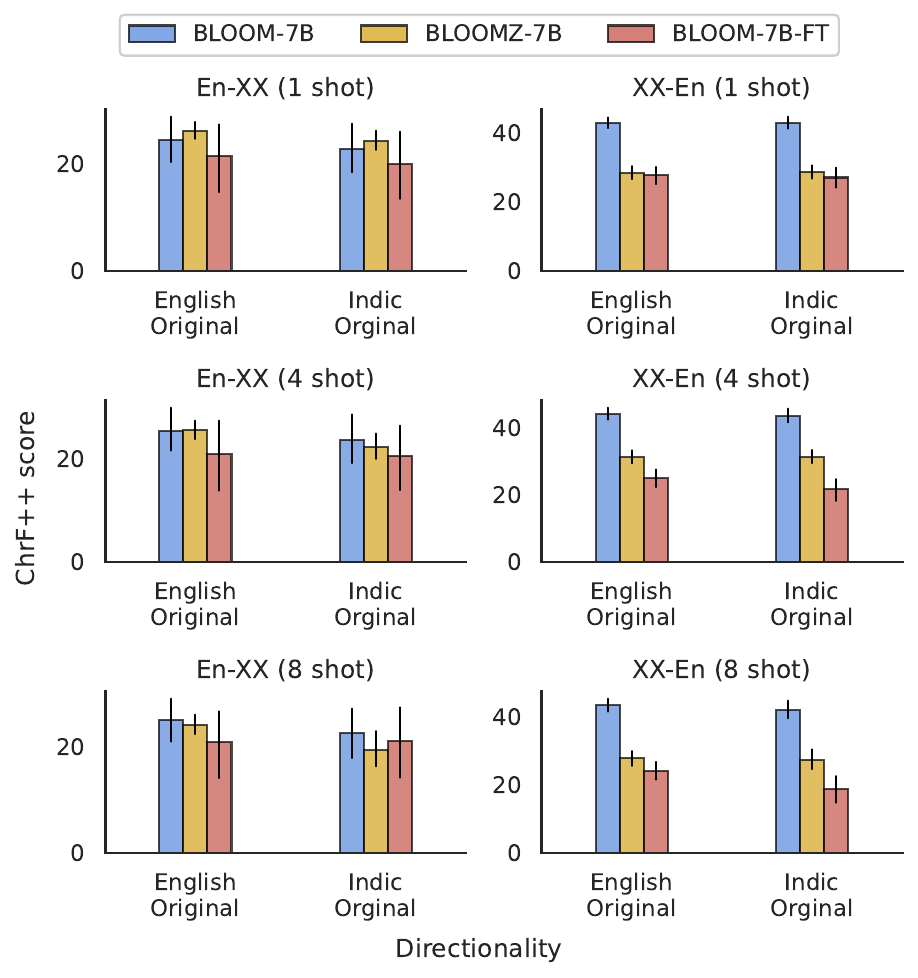}
    \caption{Shot-wise comparison of ChrF++ score of different models averaged across translation directions, comparing the choice of the source original and target original demonstrations}
    \label{fig:directionality-trends-shotwise}
\end{figure*}

\Cref{fig:instruct-trends-shotwise,fig:heterogeneous-I-trends,fig:heterogeneous-II-trends,fig:heterogeneous-III-trends,fig:heterogeneous-IV-trends,fig:implicit-trends-shotwise,fig:directionality-trends-shotwise} illustrate fine-grained details such as shot-wise and case-wise trends for the aspects outlined in \Cref{sec:methodology}. The overall trends have been described in \Cref{sec:results}.

\clearpage

\clearpage

